# DUSTrack: Semi-automated point tracking in ultrasound videos

Praneeth Namburi[1,2,†,*], Roger Pallarès-López[3,†], Jessica Rosendorf[3], Duarte Folgado[4], Brian W. Anthony[1,2,3,*]

**Affiliations**

[1] Institute for Medical Engineering and Science, MIT; Cambridge, MA 02139, USA.

[2] MIT.nano Immersion Lab, MIT; Cambridge, MA 02139, USA.

[3] Department of Mechanical Engineering, MIT; Cambridge, MA 02139, USA.

[4] Fraunhofer Portugal AICOS; Porto, 4200-135, Portugal.

† These authors contributed equally

* Corresponding authors. Email: praneeth@mit.edu; banthony@mit.edu

**Abstract**

Ultrasound technology enables safe, non-invasive imaging of dynamic tissue behavior, making it a valuable tool in medicine, biomechanics, and sports science. However, accurately tracking tissue motion in B-mode ultrasound remains challenging due to speckle noise, low edge contrast, and out-of-plane movement. These challenges complicate the task of tracking anatomical landmarks over time, which is essential for quantifying tissue dynamics in many clinical and research applications. This manuscript introduces DUSTrack (Deep learning and optical flow-based toolkit for UltraSound Tracking), a semi-automated framework for tracking arbitrary points in B-mode ultrasound videos. We combine deep learning with optical flow to deliver high-quality and robust tracking across diverse anatomical structures and motion patterns. The toolkit includes a graphical user interface that streamlines the generation of high-quality training data and supports iterative model refinement. It also implements a novel optical-flow-based filtering technique that reduces high-frequency frame-to-frame noise while preserving rapid tissue motion. DUSTrack demonstrates superior accuracy compared to contemporary zero-shot point trackers and performs on par with specialized methods, establishing its potential as a general and foundational tool for clinical and biomechanical research. We demonstrate DUSTrack's versatility through three use cases: cardiac wall motion tracking in echocardiograms, muscle deformation analysis during reaching tasks, and fascicle tracking during ankle plantarflexion. As an open-source solution, DUSTrack offers a powerful, flexible framework for point tracking to quantify tissue motion from ultrasound videos. DUSTrack is available at https://github.com/praneethnamburi/DUSTrack.

## Introduction

Tracking tissue motion—including that of skeletal muscles and organs—has clinical, engineering, and scientific applications. For example, tracking the heart wall or liver motion[1] can inform medical diagnoses, surgical planning, and disease progression monitoring. Tracking skeletal tissues can aid in identifying injury risk[2] and developing effective physical therapy and rehabilitation protocols[3]. In sports science, tracking muscle and connective tissue dynamics can help optimize athletic performance and training methods[4]. Beyond practical applications, measuring tissue motion can reveal fundamental insights into the biomechanics and neuromuscular control of human movement.

Ultrasound imaging offers unique advantages for studying tissue motion *in vivo*. Unlike magnetic resonance imaging (MRI), which restricts participants to confined spaces, ultrasound allows nearly unrestricted movement during data collection. While fluoroscopy and dynamic computed tomography (CT) can image moving body structures[5–8], their radiation exposure makes them unsuitable for non-critical applications. Other methods like sonomicrometry[9] and magnetomicrometry[10] can capture tissue motions at high spatiotemporal resolutions but require invasive procedures. Ultrasound imaging is safe, non-invasive[11], non-invasive, and provides excellent spatial (micrometer-scale) and temporal (~10–300 Hz) resolutions, making it suited for capturing dynamic tissue behaviors—from slow respiratory movements[12] to rapid cardiac or muscular activities during running[13].

Point tracking—the ability to track arbitrary points in videos—provides a foundational approach to analyzing the rich information captured in B-mode ultrasound data. Rather than building custom algorithms to measure specific features like fascicle length, muscle boundaries, or arterial diameter, a robust point-tracking system can serve as a versatile solution. By accurately following specific points of interest over time, we can derive higher-order features such as muscle area, localized torsion, and pennation angle. Therefore, point tracking sets the foundation for creating reliable, generalized tracking-based measurement frameworks.

Automated tracking in ultrasound B-mode videos remains challenging due to inherent limitations such as speckle noise[14,15], out-of-plane motion[16], and poorly defined tissue boundaries[17]. Current tracking approaches are prone to two types of inaccuracies: "drift" and "jitter." Drift refers to low-frequency tracking errors that accumulate over time, while jitter describes high-frequency frame-to-frame noise in position estimates. Traditional computer vision methods, especially those using optical flow algorithms, struggle with drift-based inaccuracies during extended tracking periods[18]. While recent deep learning approaches, especially convolutional neural network (CNN)-based trackers, often outperform traditional methods, they frequently produce jittery outputs that compromise the reliability of derived measurements like tracker velocity and acceleration[19]. While combining these approaches could potentially address their respective limitations, developing an effective integration strategy remains challenging. Recent attempts at reducing both drift and jitter have been application specific. For example, UltraTimTrack[20] excels at fascicle tracking by addressing both drift and jitter, but is not designed for the broader task of tracking arbitrary points in B-mode ultrasound videos. To address drift and jitter in the generalized context of point tracking, here we introduce DUSTrack (a **D**eep learning and optical flow based toolkit for **U**ltra**S**ound **Track**ing).

DUSTrack is a novel semi-automated, open-source toolkit for tracking arbitrary points in ultrasound videos. It employs deep learning frameworks to process spatial features within individual video frames, eliminating drift, and refines temporal trajectories using a novel optical-flow-based filter to significantly reduce jitter. An ideal point-tracking tool should excel in three key areas: generalization (across anatomical locations and imaging conditions), accuracy (precise point localization with minimal drift and jitter), and automation (minimal manual intervention). In developing DUSTrack, we prioritized generalization and accuracy to create a crucial stepping stone toward this ideal solution. The toolkit, supported by a graphical user interface, facilitates the generation of high-quality point trajectories, which can be used as ground truth data for training future fully automated point-tracking pipelines.

**Methods**

DUSTrack integrates deep learning and optical flow techniques to deliver robust point tracking in ultrasound videos. The method (Fig. 1) consists of three main components: (1) a user-friendly application to generate reliable training data for deep learning algorithms, (2) deep learning-based tracking to eliminate drift, and (3) optical flow refinement to reduce jitter in the tracking results.

The first step generates labels for fine-tuning existing deep learning models. Traditional annotation methods for natural videos typically select several frames that are highly dissimilar and require annotators to manually identify matching landmarks across disconnected frames[21]. Although effective in natural video contexts, this approach becomes impractical for ultrasound videos due to the inherent complexity of ultrasound imaging, including speckle noise[14,15], nonlinear tissue deformation[22], and out-of-plane motion[16]. These challenges make it difficult for human annotators to reliably identify corresponding anatomical points across temporally separated frames, leading to potential inaccuracies and inconsistent annotations.

In response, we adopt an alternative annotation strategy informed by common practices in ultrasound interpretation, where annotators naturally track points through sequential frames. To operationalize this intuitive approach, we developed a graphical user interface (UI) that streamlines the annotation process, allowing users to follow points through sequential frames. The UI automatically generates intermediate tracking estimates between manually annotated frames using the Lucas-Kanade[23] optical flow algorithm with reverse sigmoid tracking correction (LK-RSTC)[18], reducing manual effort while augmenting the training dataset for deep learning models. Built-in verification and correction tools enable users to review and refine these estimates, producing high-quality annotations suitable for training deep learning models. This step typically produces a few hundred labeled frames, with approximately 20 frames labeled manually (in the span of about 600 sequential frames) and the rest are augmented using the LK-RSTC algorithm. The augmentation step is optional (see results).

The second step involves training a deep learning model using annotated frames. While DUSTrack's annotations can be used with various deep learning frameworks, the toolkit is specifically designed to work with DeepLabCut (DLC)[24,25]. Through DLC, users can fine-tune state-of-the-art neural networks including ResNet[26], MobileNet[27], and EfficientNet[28]. A key advantage of these models is that they process frames independently without using temporal information, which prevents the accumulation of tracking errors (drift) commonly seen in sequential tracking methods. After training, our interface enables users to visualize DLC tracking results, make additional annotations, and iteratively refine the model. When DLC estimates are unsatisfactory, users can override them either manually or by using the LK-RSTC algorithm for improved tracking accuracy.

Finally, the optical flow refinement step employs the LK-RSTC algorithm once again to reduce jitter in the DLC estimates. While a simple low-pass filter could achieve this, utilizing the LK-RSTC algorithm better preserves fast tissue movements while minimizing jitter. The algorithm tracks points in both forward and reverse directions using the Lucas-Kanade algorithm. The final trajectory is derived by applying a weighted average of these bidirectional trajectories, with weights governed by a smooth sigmoid function that transitions preference from forward to reverse tracking across the duration of the sequence[18].

The optical flow refinement step uses a "transposed" sliding window filter. Unlike traditional sliding window filters, which estimate the signal at each point by averaging neighboring points temporally, our transposed variation averages multiple estimates derived from different overlapping windows positioned across the same frame. This approach mitigates jitter by capitalizing on the short-duration accuracy of LK-based estimations while maintaining motion characteristics.

For example, consider a 50 Hz ultrasound video analyzed using a sliding window of 0.6 seconds (30 frames). Within each window, a "tracklet" is computed, producing intermediate estimates across 28 frames using LK-RSTC. As the window moves forward, additional overlapping tracklets are computed, and the final position of any given point is the average across all tracklet-derived estimates encompassing that point. This averaging process exploits LK's established low short-term drift and jitter characteristics[23], addressing limitations found when directly interpolating between sparse deep learning predictions.

Within DUSTrack, Lucas-Kanade optical flow serves three interconnected yet distinct purposes: generating reliable reference annotations to enhance initial human labels, augmenting training datasets for deep learning models, and filtering the deep learning model outputs. Collectively, these integrated methods provide comprehensive justification for our methodological approach, supported by prior research validating these techniques in complex video tracking contexts[18,24,25].

User interface

Our interactive user interface (Fig. 2) enables users to annotate and track points in video frames. The interface streamlines manual point labeling, handles multiple annotation layers, and incorporates optical flow algorithms for interpolation and refinement of annotations.

The user interface features three main vertically stacked panels that work together to provide the necessary visualization and control for point annotation. The video display panel shows the current frame and allows users to inspect and annotate points directly on the video. Two additional trace panels display the x and y coordinates of annotations over time, providing a visual of the temporal evolution of the annotations. The visualization system displays annotations as scatter points on the video frame and as traces in the coordinate panels, with options to toggle trace visualization between dot-like and line-like display styles.

The user interface streamlines annotation through several powerful features. Users create point annotations by selecting labels and clicking locations in video frames. Annotations for tracked points are organized into layers, with each layer containing annotations for all tracked points. This allows users to maintain different versions—such as annotations created by different human annotators or deep learning models. For easy comparison, users can visualize two annotation layers simultaneously: a primary layer and a translucent overlay layer, both displayed across all three panels. Each tracked point is assigned a numeric label (e.g., "0" or "1"), and the trace panels display the trajectory of the currently selected point.

The interface features intuitive navigation controls for managing four key elements: the video frame, primary annotation layer, overlay annotation layer, and annotation label. Users can efficiently cycle through these elements using keyboard shortcuts and quickly navigate to specific frames by clicking on the trace panels. The interface clearly displays the current state of all four elements (for example: Frame 5614, primary annotation layer: labeled_data, overlay annotation layer dlc_iteration-0_100000, annotation label 8).

DUSTrack streamlines the process of creating initial training labels for deep learning models through several features that assist in producing manual annotations. When manually tracking a point, the "guess" feature uses the Lucas-Kanade algorithm to provide initial position estimates based on nearby labeled frames, which users can then refine. When tracking multiple points in a video, it can be helpful to focus on one point at a time. However, while labeling subsequent points after the first one, precisely stopping at previously annotated frames can be challenging. The UI addresses this with a "pause at next annotated frame" feature that automatically pauses at any frame containing annotations. For consistency, the toolkit includes a "trim" feature that removes incomplete frame annotations. For example, if tracking 4 points and some frames only have 3 points labeled, the trim feature will remove those incomplete frames to ensure all remaining frames have all 4 points labeled. To help users refine point locations, the UI includes features for toggling between the current frame and adjacent annotated frames.

Beyond streamlining the initial training data generation process, the UI also provides tools for generating additional training data for refining the deep learning model when model performance is poor. Users can copy annotations between layers and manage them within specific time ranges on the trace panels—including removing or copying annotations between layers. Advanced features include interpolation and refinement capabilities using the Lucas-Kanade optical flow algorithm. This allows users to automatically generate annotations between frames containing proper labels, which can be applied to an individual label or all labels in a layer.

For example, during model refinement, we often encounter cases where a trained model fails across several consecutive frames despite producing excellent estimates in surrounding frames. In such cases, users can interpolate values for the failed frames using surrounding frame data using optical flow, adjust and copy these annotations to a new layer as needed, and proceed with refining the weights of the deep learning model.

Each annotation layer can be saved to a JSON file, which includes frame numbers and 2D coordinates for each label. These files are automatically loaded when returning to a previously annotated video, allowing for seamless continuation of work across sessions.

DUSTrack's UI leverages several open-source Python libraries to create an intuitive interface that can be easily customized and extended. The datanavigator library—a thin wrapper around the popular matplotlib library created by the authors of DUSTrack—creates the UI and visualizations. Event-driven programming in the datanavigator module enable interactive features like annotation placement and navigation. The pysampled package, also developed by the authors, handles time series data using NumPy and SciPy under the hood. The pandas library primarily ports annotations to and from DeepLabCut. The decord library reads videos, while OpenCV provides the implementation of the Lucas-Kanade optical flow algorithm.

Datasets

To evaluate DUSTrack and demonstrate its applications, we used three datasets. First, we used an internal dataset of transverse B-mode ultrasound videos capturing the upper arm during a reaching task in 36 healthy adults[29]. These videos contain approximately 11 annotated points tracking the triceps, brachialis, and humerus bone to analyze musculoskeletal tissue dynamics. Second, we assessed DUSTrack's fascicle tracking capabilities using a public dataset specifically designed for medial gastrocnemius tracking[30]. Finally, we demonstrated cardiac point-tracking applications using a video from the EchoNet-LVH dataset[31].

Statistical Analysis

Statistical analyses were conducted using the Scipy and statsmodels Python libraries. Scipy was used to perform paired t-tests for pairwise comparisons between DUSTrack and UltraTimTrack in fascicle tracking, as well as between the fine-tuned ResNet-50 and zero-shot models. For the latter comparisons, statsmodels was also used to apply a Bonferroni correction for multiple testing. Additionally, Scipy was employed for performing a binomial test to assess whether LK-RSTC label augmentation yielded superior tracking performance based on the blinded evaluator's ratings.

**Results**

State-of-the-art automated point tracking models demonstrate limited success when applied to ultrasound imaging

Recent advances in automated point tracking, particularly deep learning-based Track-Any-Point (TAP) methods[32–35], have achieved impressive performance on natural (RGB) videos. These models enable zero-shot tracking: given a point annotation in a single frame, they track this landmark across the entire video. State-of-the-art TAP models have achieved impressive performance in natural (RGB) videos, demonstrating robustness even under prolonged occlusions.

Motivated by the success of TAP models in videos of natural scenes, we investigate whether they can be directly applied to track points in B-mode ultrasound videos. We compare the performance of these zero-shot tracking models to the performance of a ResNet-50 model finetuned to each video. The ground truth was DUSTrack-assisted human annotations that were refined over several (typically 2-4) iterations and manually corrected.

To assess performance and generalization across varied tracking approaches, we survey state-of-the-art TAP models and selected four—CoTracker3[32], BootsTAP[33], LocoTrack[34], and PIPs++[35]. Our selection is based on three key factors: (1) reported performance on TAP-Vid and real-world video datasets, (2) architectural diversity driven by tracking strategies, both context-aware (CoTracker3, BootsTAP, which refine tracking using surrounding trackers) and agnostic

(LocoTrack, PIPs++), and (3) their use of self-supervised learning techniques to improve generalization and reduce the need for annotated data (CoTracker3 and BootsTAP).

Our results show that ResNet-50 models fine-tuned with 25 human-annotated frames significantly outperform state-of-the-art TAP models in tracking accuracy (Fig. 3a). This demonstrates a clear gap between finetuned and zero-shot approaches. Power spectral analysis of point trajectories shows varying levels of high-frequency components, revealing model-specific differences in frame-to-frame jitter (Fig. 3b,c). Notably, CoTracker3 achieves jitter levels closest to the ground truth, even slightly outperforming the supervised ResNet-50 (Fig. 3b,c). This indicates that while CoTracker3 effectively suppresses high-frequency jitter, it remains less precise in localizing the tracked point over time (Fig. 3a-c). Analysis of trajectories over several seconds (Fig.3c) highlights these distinct differences in jitter levels across models while emphasizing the performance gap between finetuned and zero-shot tracking methods, with CoTracker3 emerging as the current leader in zero-shot tracking for ultrasound applications (Fig.3a).

While current state-of-the-art zero-shot tracking solutions have largely mitigated drift, they still need significant improvements to match human-level accuracy and reduce frame-to-frame jitter in ultrasound videos. To address the issue of frame-to-frame jitter, we introduce a filtering method that preserves fast tissue motions while reducing jitter.

### LK-RSTC filtering effectively reduces tracking noise while preserving high-frequency motion

The DUSTrack workflow introduces a post-processing filtering step, based on the LK-RSTC algorithm[18], to leverage the complementary strengths of deep learning and optical flow approaches (Fig. 4a-d). While deep learning models excel at identifying points in individual frames, they often introduce frame-to-frame jitter that can obscure underlying motion patterns. In contrast, optical flow algorithms provide smooth tracking over short durations but accumulate errors over longer periods. Our solution creates multiple short-duration "tracklets" using the LK-RSTC algorithm, with deep learning estimates serving as anchor points (Fig. 4a). By sliding this window frame-by-frame and averaging overlapping tracklet estimates (Fig. 4b,c), we achieve smooth, physically plausible trajectories while maintaining the global accuracy provided by deep learning models (Fig. 4d).

Analysis of frequency spectra reveals the limitations of traditional low-pass filtering approaches. The LK-RSTC algorithm produces traces and frequency characteristics that closely match the ground truth signal (Fig. 4e,f). While a 10 Hz low-pass filter fails to adequately suppress high-frequency noise, a more aggressive 5 Hz cutoff introduces excessive signal attenuation that distorts the underlying motion patterns (Fig 4g,h). Representative traces demonstrate this effect qualitatively: low-pass filtered trajectories either show spurious motions or remove faster motion components, whereas the LK-RSTC approach better captures true motion patterns while effectively removing noise (Fig. 4f,h,i).

### LK-RSTC Label Augmentation: Benefits and Limitations

LK-RSTC label augmentation is an optional component of DUSTrack for augmenting manual annotations used to finetune the ResNet-50 models. To determine its efficacy, we compare the performance of ResNet-50 models with 25 human-annotated frames (across ~500 frames), with and without augmenting the training dataset with predictions in the intermediate frames generated using the LK-RSTC algorithm. Augmenting the training dataset reduces jitter in model outputs (Fig. 5a,b), making model refinement easier.

Although augmentation reduces jitter and improves accuracy, upon applying the LK-RSTC filter, the power spectra of trajectories with and without the LK-RSTC label augmentation step are nearly identical (Fig. 5e). However, the results from the two approaches are significantly different from each other (Fig. 5c,d). To better understand the efficacy of the augmentation step, we conduct a systematic evaluation to determine whether any performance differences between the two approaches—with and without augmentation—are perceptible to a human evaluator after applying the LK-RSTC filter. Using 388 point trajectories from 36 videos (one per participant), we overlay both outputs on each video segment for comparison. A blinded evaluator, unaware of which algorithm produced which output, indicates their preference between the first output, second output, or both. The results do not indicate a systematic preference for LK-RSTC augmentation: 152 trajectories receive no preference, 127 trajectories with LK-RSTC augmentation are preferred, and 109 trajectories without LK-RSTC augmentation are preferred (Fig. 5e). These findings indicate that the impact of LK-RSTC augmentation becomes imperceptible after applying the final LK-RSTC filtering step. Nonetheless, the LK-RSTC label augmentation tool may still serve a valuable role during annotation by providing visual feedback, allowing users to verify label quality based on the coherence of interpolated motion.

Based on the data from our blind evaluation experiment, we estimate the perceptual threshold at which humans can no longer detect differences between tracked trajectories. To estimate this threshold, we calculated two values: the average difference between outputs when evaluators could not distinguish between them (90 μm), and the average difference when evaluators could identify a preferred output (140 μm) (Fig. 5d). We propose that the perceptual threshold for our imaging conditions lies between these two values, at approximately 100 μm.

Choosing the number of annotated frames

We then investigate the relationship between tracking accuracy and the number of annotated frames in videos with cyclic motion patterns, such as cardiac cycles or walking. Annotating frames within 1-2 motion cycles provides an effective basis for training because human annotators are also prone to introducing drift errors. As expected, increasing the number of manual annotations improves tracking accuracy (Fig 6). We observe a sharp error reduction when increasing annotations from 5 to 15 frames, followed by diminishing returns beyond this point. Based on an accuracy-effort trade-off argument, we recommend annotating approximately 25 frames within 1-2 motion cycles for optimal performance.

Applications of DUSTrack in Clinical and Biomechanical Analysis

Ultrasound imaging is widely used in medical and biomechanical applications to measure important physiological changes, such as the thickness variations of the heart's interventricular septum during cardiac cycles or muscle fascicle length changes during walking. These valuable measurements can be obtained through point tracking techniques. To showcase how point tracking can be effectively leveraged to extract these measurements from B-mode ultrasound videos, we apply DUSTrack across three distinct tracking tasks that encompass both clinical and biomechanical applications.

First, we extract dynamic measurements from parasternal long-axis echocardiograms in the EchoNet-LVH dataset[31]. We measured the interventricular septum (IVS) thickness, left ventricular internal diameter (LVID), and left ventricular posterior wall (LVPW) thickness over time. These measurements are essential for assessing cardiac function and diagnosing conditions like left ventricular hypertrophy (LVH), which increases cardiovascular event risk[36,37]. We tracked points on both sides of the IVS and LVPW using the DUSTrack workflow across four cardiac cycles (Fig. 7a,b).

Next, we apply the point tracking framework to analyze muscle deformations in the upper arm during a reaching task[38]. In a cross section showing the triceps and brachialis muscles, we track two points in the superior-inferior (vertical) direction and two points in the medial-lateral (horizontal) direction in each muscle to quantify deformations during the reaching movement (Fig. 7c). We track these deformations over time (Fig. 7d) by computing them as $(L - L_0)/L_0$, where $L_0$ represents the initial length at the start of the task. This analysis reveals distinct behaviors in both muscles: the brachialis shows correlated horizontal and vertical deformations, indicating changes in cross-sectional area, while the triceps exhibits anti-correlated deformations, reflecting changes in muscle shape.

Finally, we apply the DUSTrack framework to fascicle tracking—a well-studied problem in biomechanics. Modern fascicle tracking algorithms effectively extract fascicle length and pennation angle using classical computer vision algorithms, often achieving excellent results through fully automated processes. In the gastrocnemius muscles, fascicle tracking requires identifying both aponeurosis positions and fascicle angles throughout the ultrasound video. We define the upper aponeurosis line by labeling two points, then mark two points on the lower aponeurosis, with one point marking where it intersects a fascicle. We add another point along this fascicle to define the fascicle orientation. Using geometric calculations, we compute the fascicle length (the distance between aponeurosis intersections) and the pennation angle (the angle between the fascicle and lower aponeurosis) (Fig. 7e). By calculating these metrics frame by frame, we track changes in fascicle length and pennation angle across multiple cycles of ankle dorsiflexion and plantarflexion (Fig. 7f).

We evaluated DUSTrack's performance against state-of-the-art fascicle tracking algorithms (Fig. 8). For comparison, we use a public dataset[30] containing recordings from 5 participants—with annotations from 3 expert raters across 3 separate days—totaling 9 manual annotation for each of 18 annotated frames per participant. The average across all 9 ratings is considered ground truth. We compare DUSTrack against three existing approaches: UltraTimTrack[20], HybridTrack[39], and DLTrack[40]. Due to severe failure cases (extremely poor tracking performance or failure to process the video), we exclude results from HybridTrack and DLTrack, limiting our quantitative comparisons to UltraTimTrack. In measuring fascicle length, DUSTrack shows accuracy comparable to UltraTimTrack (Fig. 8c), with DUSTrack slightly underestimating and UltraTimTrack slightly overestimating lengths (Fig. 8d). Both methods show comparable accuracy for estimates of pennation angle, with both methods slightly underestimating the angle compared to ground truth (Fig. 8ef).

Together, these applications highlight the versatility of DUSTrack-based point tracking for both clinical and biomechanical use cases, tracking various parameters of interest, and enabling interpretable measurement of diverse tissue dynamics from B-mode ultrasound videos.

**Discussion**

A General-Purpose Tracking Solution for Ultrasound

Despite ultrasound's widespread use in clinical and research settings, most landmark tracking methods are tailored to specific anatomical structures, limiting their adaptability across different regions, orientations, probe configurations, and research applications. In contrast, the ability to track arbitrary points across an ultrasound video enables more flexible and exploratory analyses. Whether monitoring tissue displacement, tracing a vessel wall, examining fascicle orientation, or following a surgical tool, a generalized tracking framework allows researchers and clinicians to extract features most relevant to their application, without the need to develop custom tools for each new task.

One widely used technique for measuring soft tissue deformation is speckle-trackig[41], particularly in clinical assessments of cardiac function[42]. It employs block-matching or cross-correlation techniques to track the displacement of speckle patterns between frames. However, conventional speckle tracking techniques are application specific and to generalize, typically require tuning with gold-standard reference data obtained through methods such as sonomicrometry, which can be expensive and impractical[43,44].

In a different domain, fascicle tracking is widely used in biomechanics to measure fascicle length and pennation angle, especially in the medial gastrocnemius. The consistent anatomy of this muscle allows traditional computer vision methods – such as line detection and Hough transforms[20,39] – to fit lines through tracked points along the fascicles and aponeuroses. Yet, these are limited to muscles with specific anatomy: two distinct aponeuroses and a steep pennation angle that brings fascicle intersections into view. Even small anatomical changes, such as switching to the lateral gastrocnemius, can significantly reduce accuracy.

DUSTrack offers a promising alternative to task-specific tracking methods, serving as a flexible and general-purpose framework for point tracking in ultrasound. It represents a key stepping stone toward the development of zero-shot ultrasound tracking algorithms. As point tracking methods continue to improve in generalization and accuracy, we anticipate they will ultimately replace conventional tailored algorithms by delivering high-quality measurements with minimal manual intervention.

Achieving High-Quality Tracking Still Requires Manual Intervention

The ideal point-tracking tool would offer broad generalization, high accuracy, and minimal manual intervention. DUSTrack takes a meaningful step toward this goal by prioritizing generalization and accuracy across diverse ultrasound applications. However, achieving fully automated, zero-shot tracking in ultrasound remains an open challenge.

Foundation models for point tracking are already being developed in the context of natural videos, enabling zero-shot tracking. Yet, no general-purpose model currently exists for ultrasound that can match DUSTrack's accuracy, required for medical use. Solutions such as EchoTracker[45] remain specialized for echocardiograms, while PIPsUS[46] requires self-supervised training for each new task and still struggles to achieve sufficiently high accuracy.

Reaching this level of automation, while meeting the high accuracy demands of medical imaging, will likely require large-scale, diverse datasets that enable models to generalize across anatomical regions, imaging conditions, and clinical tasks. DUSTrack is well positioned to contribute to this next phase by providing an efficient, user-guided workflow for generating high-quality point tracking data at scale. Grounded in a data-centric AI perspective[47,48], DUSTrack prioritizes high-quality, context-aligned annotations, achieving strong performance with approximately 25 well-placed labels. Its continued application across varied use cases will be essential for building the training resources needed to develop future models that move closer to the ideal of zero-shot ultrasound tracking.

Unlike standard computer vision workflows that annotate spatially diverse, disconnected frames, DUSTrack adopts a sequential annotation strategy tailored to the visual complexity of ultrasound imaging. In this context, certain features (such as within elastic tissues) are more reliably tracked when annotators inspect contiguous frames. This approach improves label consistency, enables effective augmentation through optical flow-based interpolation, and reduces overall manual effort.

A key consideration in evaluating tracking performance is the perceptual threshold—the smallest displacement that a human annotator can reliably detect in tracked trajectories in ultrasound videos. Based on our data collection procedures and the annotators involved, we estimate this threshold to be approximately 100 μm (Fig. 5d). The accuracy of a finetuned ResNet-50 model—the first step in the DUSTrack workflow—at approximately 200 μm (Fig. 3a), approaches this threshold, indicating that its performance is nearing both the perceptual and physical resolution limits of ultrasound imaging. This threshold also serves as a practical benchmark for zero-shot tracking models, which currently fall short of this level of precision. As such, it helps to quantify the current gap between supervised frameworks like DUSTrack and general-purpose, zero-shot models still under development.

A Modular Framework for Customizable Ultrasound Tracking

Beyond its performance and generalization capabilities, a key strength of DUSTrack lies in its modular architecture. Each component of the framework – annotation, model tracking, and post-processing – can be used independently, extended for other tasks, or replaced with alternative tools as needed. This flexibility allows DUSTrack to serve not only as a complete tracking solution but also as a customizable platform for developing and benchmarking new ultrasound analysis methods.

A central component of the framework is the graphical user interface (UI), designed for intuitive annotation and iterative refinement. It enables users to generate high-quality training data and update models efficiently. Importantly, the UI also holds standalone value: it supports the creation of task-specific datasets for adapting tracking models to new anatomical regions, imaging modalities, or research applications. In addition, it facilitates post hoc refinement by helping users identify and correct local tracking errors, allowing targeted fine-tuning to improve model accuracy. As such, the UI serves as a versatile tool for both dataset development and model validation.

Another key component is the LK-RSTC filtering step, which leverages optical flow to reduce tracking jitter while preserving the true underlying motion captured in the video. Unlike traditional low-pass filters – which may suppress physiologically meaningful high-frequency dynamics – this method uses local frame-to-frame motion estimates to selectively reduce spurious fluctuations without distorting signal. The result is a smoother trajectory that remains faithful to the original tissue dynamics, contributing to the overall goal of achieving high-accuracy, low-jitter tracking suitable for downstream quantitative analysis and extraction of higher-level metrics.

While this filtering approach is computationally efficient and sufficiently robust for many applications, it does not fully account for the complexities of cardiac or elastic tissue motion, nor does it explicitly handle imaging artifacts, motion discontinuities, or sparse motion structures. More advanced optical flow frameworks, such as those that incorporate robust M-estimators, motion sparsity constraints, or learned cardiac motion dictionaries, have shown improved performance in ultrasound motion estimation tasks[49]. Incorporating these more sophisticated techniques into DUSTrack's filtering pipeline could offer enhanced tracking stability and noise rejection, particularly in regions affected by speckle drop-out, shadowing, or reverberation artifacts. This represents a promising direction for future refinement of the toolkit.

For the core tracking model, we adopted a ResNet-50[26] architecture from DeepLabCut[24,25] as the deep learning model for point tracking. Arguably, fine-tuning a video tracking model such as CoTracker3[32], which was the best-performing zero-shot model in our evaluation, would be a more straightforward choice. While fine-tuning CoTracker3 could potentially surpass the ResNet-50 performance on ultrasound videos, these models typically require significantly larger computational resources and memory during training, which may limit their practical adoption in medical research settings where access to high-end hardware is not guaranteed. In contrast, DUSTrack is designed to be lightweight, accessible, and trainable on widely available GPUs with 8 GB of memory. We therefore prioritized an approach that is both computationally efficient and adaptable to typical medical imaging workflows, where ease of use and practical model training times are important.

Broader Clinical and Scientific Impact of General-Purpose Ultrasound Tracking

The development of a general-purpose ultrasound point tracking framework provides an adaptable methodology that can enhance clinical and biomechanical analyses, offering a complementary or improved alternative to existing measurement practices. In cardiac imaging, conventional echocardiographic workflows often rely on static measurements, typically taken via manual caliper placement at end-diastole and end-systole[50]. By guiding label placement, the DUSTrack UI may reduce manual measurement variability and promote greater inter-operator consistency, facilitating standardized assessments. Additionally, DUSTrack enables dynamic analysis throughout the entire cardiac cycle, allowing users to extract time-varying metrics such as wall thickness, chamber dimensions, and valve motion. These dynamic measurements may provide diagnostic insights beyond what static estimates can capture. For instance, tracking temporal changes in left ventricular wall thickness, chamber expansion, or interventricular septum displacement may improve evaluation of conditions such as hypertrophic or dilated cardiomyopathy[51,52].

In exercise science, movement analysis primarily focuses on the gross motions of body segments, but the motion of elastic tissues (such as skin, muscle, and internal organs) also carries substantial clinical and scientific relevance. Understanding how these tissues deform and interact during movement provides insights that are less accessible through traditional, joint kinematic analysis alone[38]. Unlike static measurements, which may be sufficient for evaluating certain features like muscle atrophy[53], dynamic measurements reveal time-dependent behaviors such as strain patterns, tissue velocity, and transient shape changes that could be essential for assessing function, efficiency, and coordination. A persistent challenge in biomechanics has been the lack of non-invasive, dynamic tools to capture these tissue-level motions in vivo. DUSTrack helps address this gap by providing a generalized accessible framework for tracking tissue deformation. The dynamic measurements derived from its point tracking can inform rehabilitation protocols, optimize sports performance, and support injury prevention strategies by offering a more complete view of muscle function during movement.

In addition to the fields of cardiology and biomechanics, we imagine the development of improved ultrasound tracking algorithms will encourage increased adoption of ultrasound by other medical and scientific fields. Since ultrasound provides high frame rate, non-invasive, non-irradiating internal images, we imagine a much broader range of applications are possible once feature extraction is simplified. Some examples include imaging the liver or kidneys, machine parts, and aiding in developing 3D volume reconstruction algorithms.

Limitations

Despite the strengths of our methodology, several limitations remain. First, DUSTrack's accuracy is fundamentally constrained by the quality and consistency of manual annotations. As discussed earlier, both the perceptual threshold of human observers and variability in annotator expertise or intra-session consistency can affect tracking performance. This highlights the need for systematic annotation protocols and potentially consensus-based labeling strategies to improve standardization.

Second, the performance of DUSTrack's optical-flow-based refinement depends on video frame rate. Lower temporal resolution can impair its ability to capture rapid physiological motion, diminishing the effectiveness of jitter reduction. Future work should evaluate the toolkit across a wider range of frame rates and imaging conditions to ensure robustness under varied acquisition settings.

Third, while the framework is semi-automated, it still requires manual input during the initial annotation and model refinement stages. Fully automating these steps – through large-scale datasets and advanced self-supervised or weakly supervised learning methods – represents a critical direction for future development.

Finally, although DUSTrack demonstrates generalization across multiple anatomical regions, its performance in clinical populations remains to be validated. Pathological cases or atypical tissue properties not represented in our dataset may pose additional challenges. Evaluating the toolkit's robustness in these contexts will be essential for broadening its clinical utility.

**Conclusion**

We introduced DUSTrack, a flexible, semi-automated framework for accurate point tracking in B-mode ultrasound videos, combining deep learning with optical flow to address key challenges in ultrasound tracking. Through its modular design, intuitive user interface, and novel filtering techniques, DUSTrack supports high-quality tracking across diverse anatomical structures and motion types. It demonstrates broad applicability and performance comparable to specialized methods, while remaining lightweight and accessible. As an open-source tool, DUSTrack provides a foundation for tissue motion quantification, scalable data generation, and the broader adoption of automated ultrasound analysis in clinical and research settings.

**Acknowledgments**
This work was carried out in part through the use of MIT.nano Immersion Lab's facilities. This material is based upon work supported by the National Science Foundation Graduate Research Fellowship Program under Grant No. 2141064. We thank Micha Feigin-Almon for configuring the ultrasound system. We thank Mariia Smyk, Andrea Leang, Kelly Wu, and Isabel Waitz for assistance with data collection and annotation. We thank Vincent Chen and Charles Williams for their assistance with data annotation. We thank all the participants for volunteering their time. Professional editing services were not used in the preparation of this manuscript. Large language models were used to make minor improvements to the writing, and were not used to generate any of the ideas presented in this work.

**Funding:**
Sekisui House (BA)
NCSOFT (BA)
"la Caixa" Foundation (ID 100010434) fellowship LCF/BQ/EU22/11930097 (RPL)
Fulbright U.S. Student Program and Fulbright Commission Portugal (DF)
European Health and Digital Executive Agency (ID 101136376) (DF)
National Science Foundation Graduate Research Fellowship (Grant No. 2141064) (JR)
Bose Fellows Program, MIT (PN)

**Author contributions:**
Conceptualization: PN
Investigation: PN, RPL, JR
Software: PN
Data analysis: PN, RPL, JR, DF
Visualization: PN, RPL, JR
Funding acquisition: BA
Project administration: PN, BA
Supervision: PN, BA
Writing – original draft: PN, RPL, JR, DF
Writing – review & editing: PN, RPL, JR, DF, BA

**Competing interests:** Authors declare that they have no competing interests.

**Data and materials availability:** The data used in this manuscript will be made publicly available through a research dataset journal. Until then, the data supporting this study's findings are available upon reasonable request from the corresponding author. DUSTrack is publicly available at https://github.com/praneethnamburi/DUSTrack. All other data processing code is available from the corresponding author upon reasonable request.

**Figures**

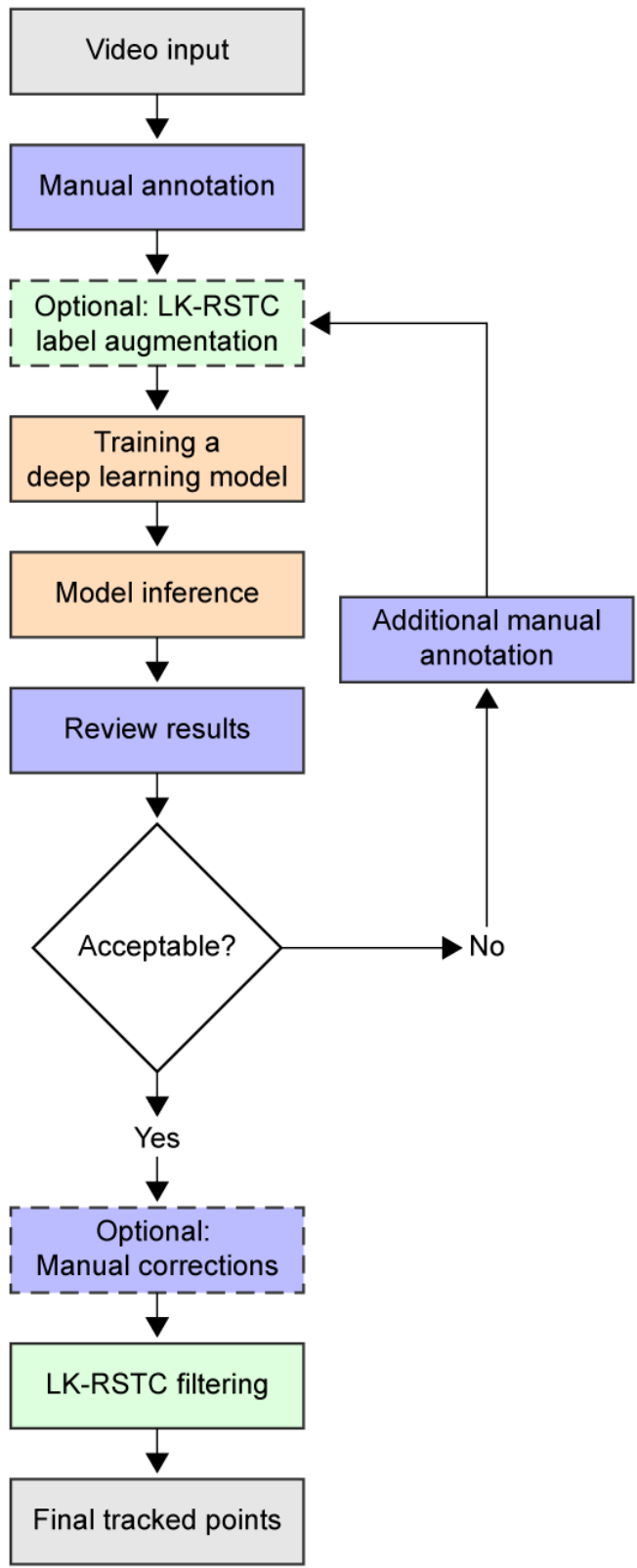

**Figure 1: DUSTrack workflow.** The process starts with video input and manual annotation of key frames. These annotations can optionally be augmented using the LK-RSTC optical flow algorithm (dashed box). The annotations are used to train a deep learning model, such as the ResNet-50. After model inference, results undergo review. If the results are unsatisfactory, users can make additional manual annotations to refine the model through iterative training. Once results meet quality standards, optional final manual corrections can be made before applying LK-RSTC filtering to produce the final tracked points. Purple indicates manual steps, light orange indicates deep learning steps, and light green indicates optical flow steps.

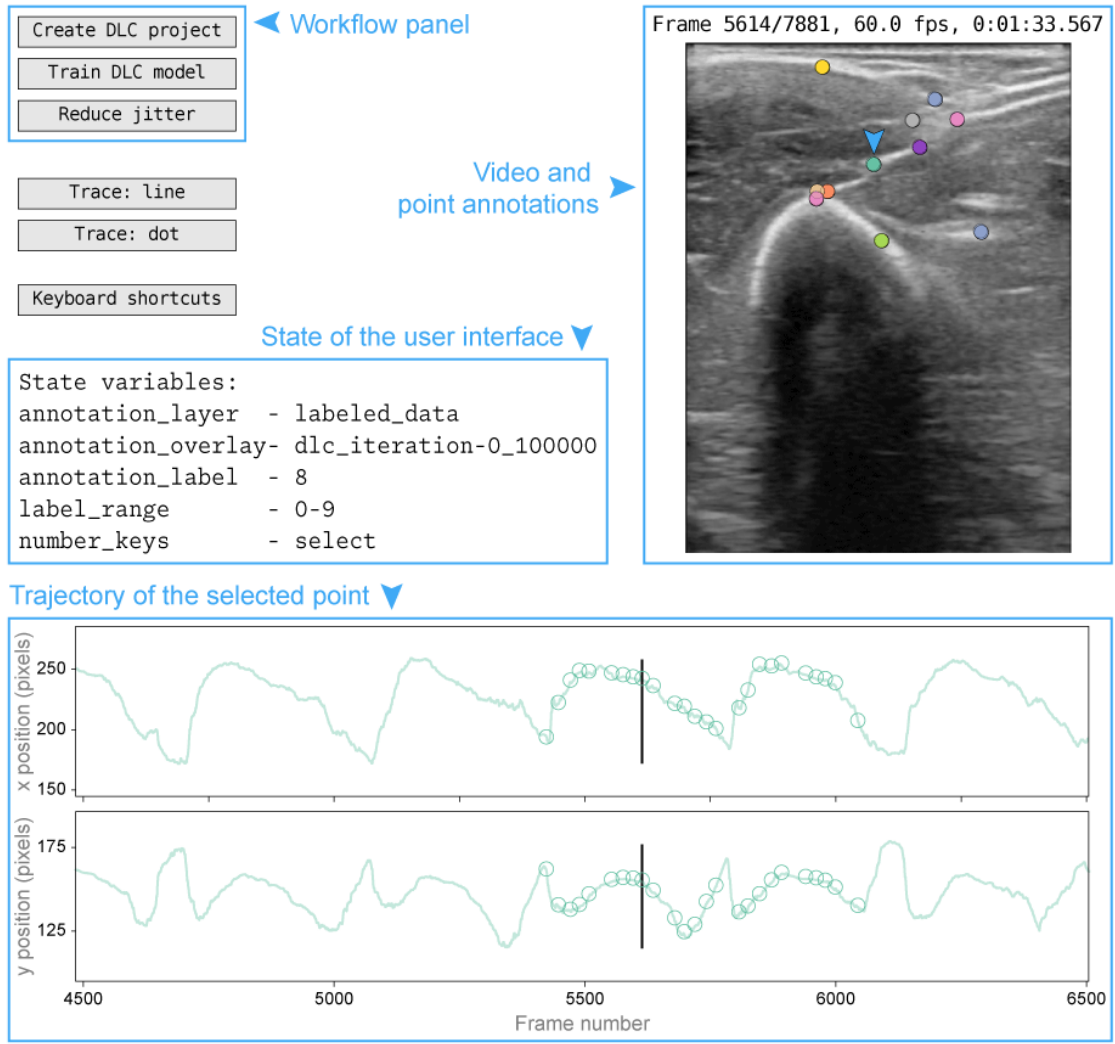

**Figure 2: DUSTrack's graphical user interface.** The workflow panel (top left) provides controls for creating DLC projects, training models, and reducing jitter. The video and point annotations panel (right) shows the ultrasound video frame with current point annotations. The selected point is marked here with an arrow. The interface state panel (middle left) displays active variables, including the current annotation layer and selected point. In the bottom is a trajectory panel with plots showing x and y coordinates of the selected point over time. The interface can display two annotation "layers" simultaneously—in this example, the "labeled_data" layer appears as open circles in the trajectory panel, while the "dlc_iteration-0_100000" layer appears as a translucent continuous trace. Blue text, arrows, and boxes have been added to highlight the interface features and are not part of the actual interface.

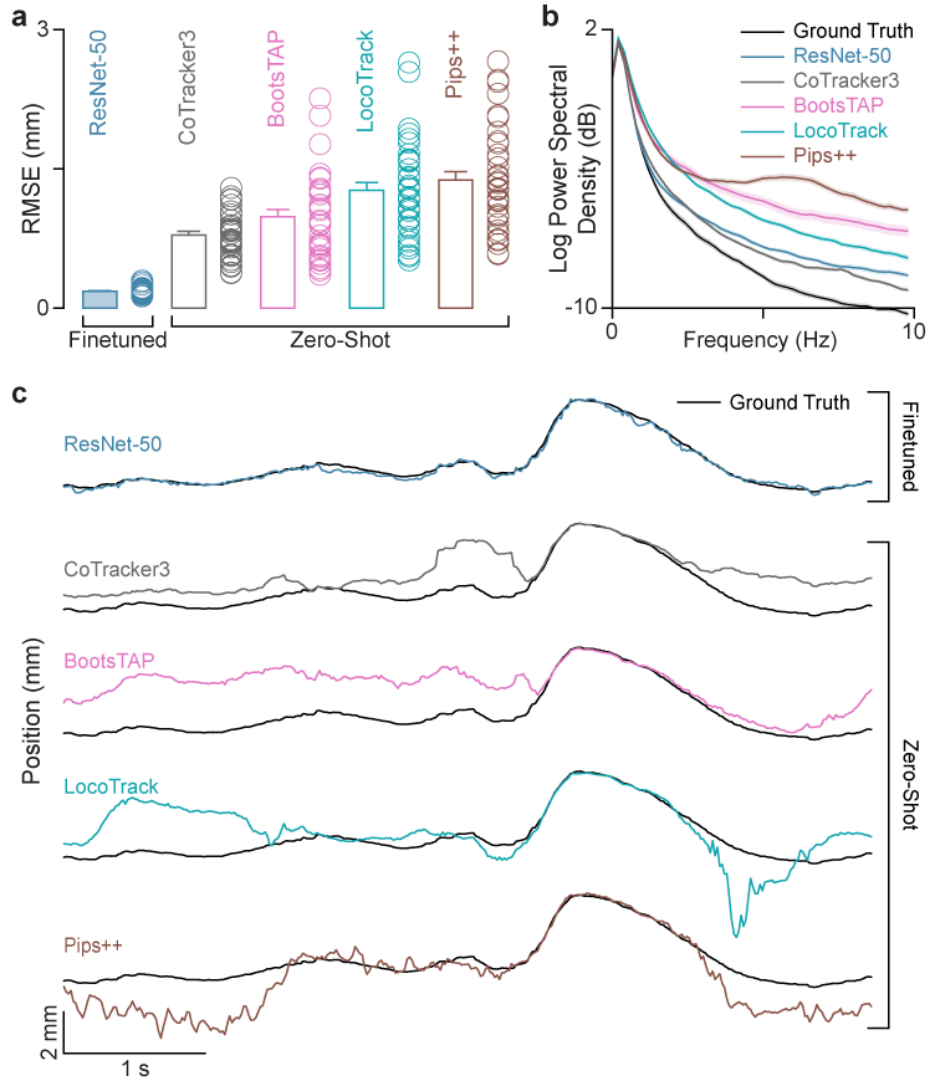

**Figure 3. A finetuned ResNet-50 tracks points in b-mode ultrasound videos with higher accuracy than current state-of-the-art zero-shot methods.**

(a) Root mean square error (RMSE) in position between the ground truth and the outputs of a finetuned ResNet-50 (trained on 25 labeled frames) and four zero-shot tracking models. The ResNet-50 has significantly lower tracking error compared to CoTracker3 (paired t-test, $t_{35}$ = -15.56, p = $2.71 \cdot 10^{-16}$), BootsTAP (paired t-test, $t_{35}$ = -10.68, p = $1.49 \cdot 10^{-11}$), LocoTrack (paired t-test, $t_{35}$ = -12.93, p = $6.84 \cdot 10^{-14}$), and Pips++ (paired t-test, $t_{35}$ = -13.30, p = $3.02 \cdot 10^{-14}$).
(b) Power spectral density of model output for the finetuned ResNet-50 and the zero-shot models. All models present higher frequency components compare to the ground truth.
(c) Representative traces of tracked point position from finetuned ResNet-50 model and zero-shot models relative to ground truth. While the fine-tuned model accurately follows the true point trajectory, the zero-shot models present noticeable tracking errors.

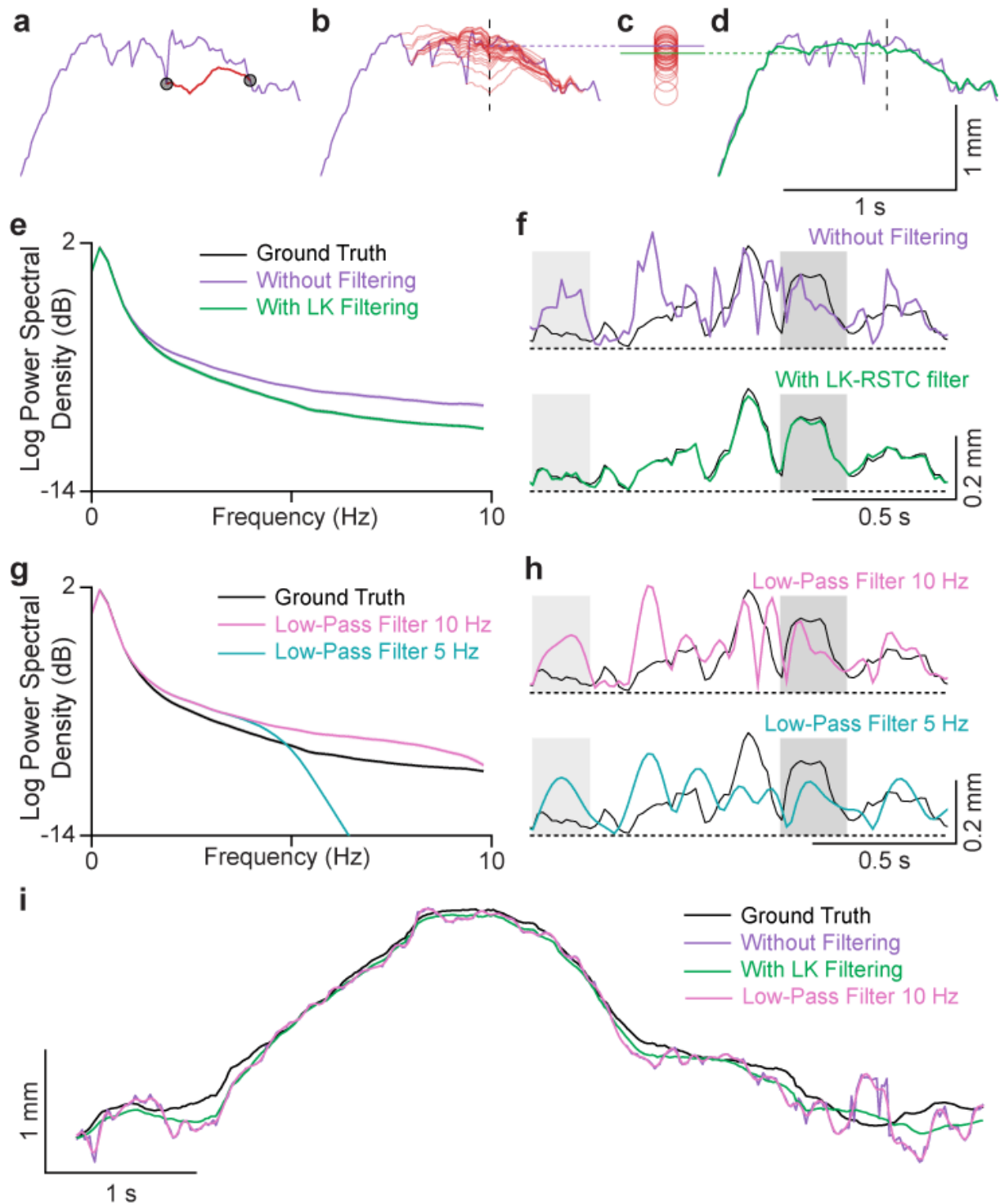

**Figure 4: The LK-RSTC postprocessing filter in DUSTrack reduces temporal jitter while better preserving both slow and fast motion dynamics compared to low-pass filtering.**

(a-d) Illustration of the LK-RSTC filtering algorithm, applied to the output of a finetuned ResNet-50 model.

(a) Model output (purple) and a tracket (red) computed using the LK-RSTC algorithm between two points (black circles).

(b) Visualization of overlapping tracklets (red translucent traces) that contribute to determining the filtered value at the dashed vertical line.

(c) The value of each tracklet at the dashed vertical line in panel b, along with the model output (purple line) and the filtered value (green).

(d) The original model output (purple) and the result after LK-RSTC filtering (green).

(e) Power spectral density of the ground truth (black), model output (purple), and LK-RSTC filtered model output (green), averaged across data from 36 participants. The black trace is not visible due to significant overlap with the green trace, and the standard error of the mean (SEM) is too small to be visible.

(f) Representative point trajectories showing ground truth (black), model output (purple), and LK-RSTC filtered model output (green). The filtered output effectively suppresses high-frequency noise (light gray boxes) while preserving genuine high-frequency motion components (dark gray boxes). All traces are high-pass filtered at 1.5 Hz to highlight differences between filtering methods.

(g) Power spectral density of the model output with 10 Hz (pink) and 5 Hz (cyan) low-pass filters. The 5 Hz filter excessively attenuates meaningful high-frequency motion, while the 10 Hz filter inadequately suppresses noise in the 2–10 Hz range.

(h) Representative point trajectories comparing ground truth (black) with 10 Hz (pink) and 5 Hz (blue) low-pass filtered outputs. The 5 Hz filter suppresses high-frequency tissue motions (dark gray box).

(i) Representative trajectories of a tracked point comparing LK-RSTC filtering against different low-pass filter settings. Unlike the traces in panels f and h, these traces are not high-pass filtered. While low-pass filters smooth the signal, they fail to eliminate model errors and suppress genuine motion dynamics. The LK-RSTC filter achieves a better balance, reducing spurious jitter while preserving meaningful high-frequency motion.

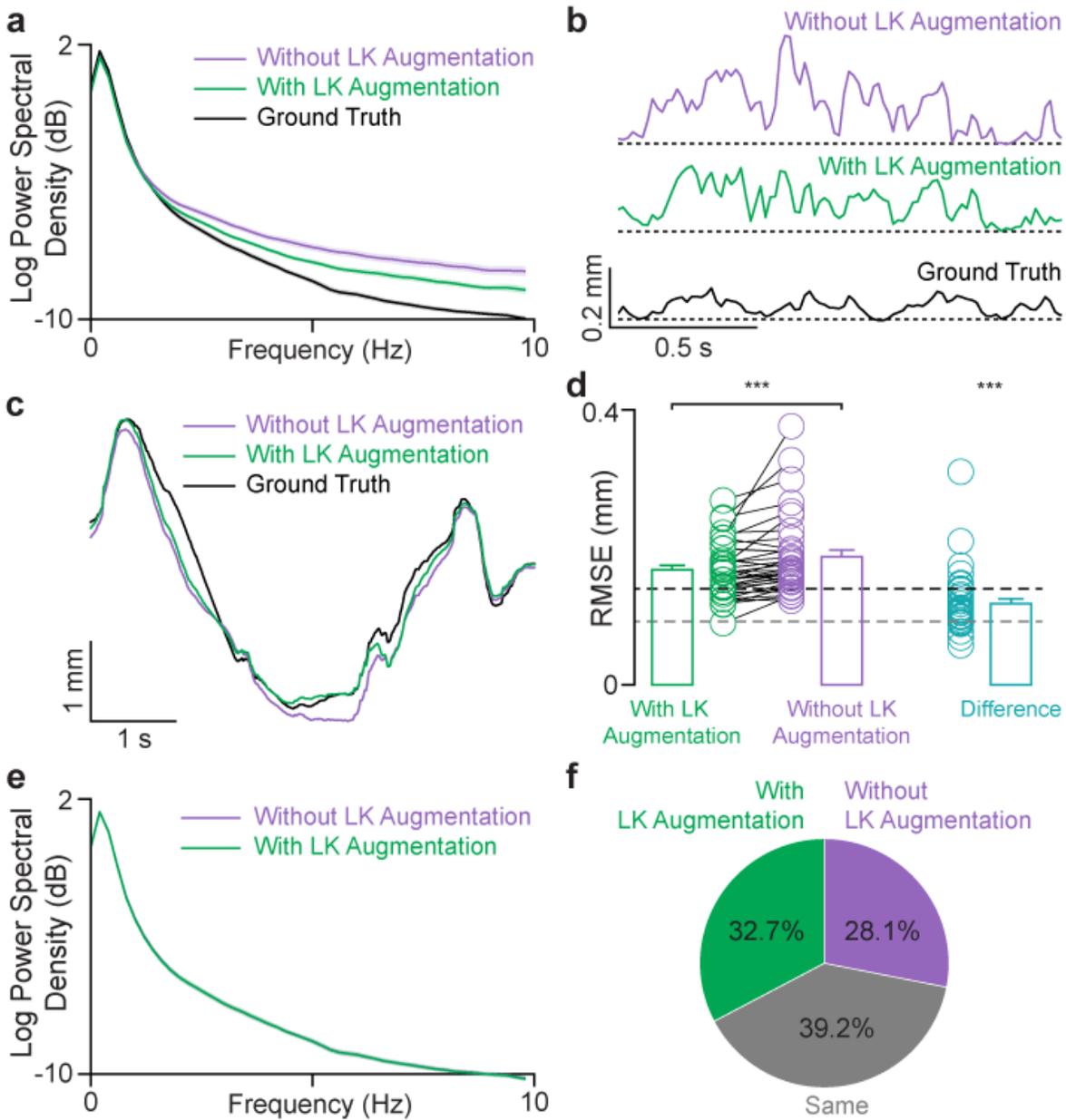

**Figure 5. The LK-RSTC label augmentation step in DUSTrack reduces jitter in initial model outputs but this effect is muted after applying an LK-RSTC filter.**

(a) The LK-RSTC label augmentation step occurs prior to finetuning the a ResNet-50 model. Power spectral density analysis of the ResNet-50 model outputs shows that models trained with LK-RSTC label augmentation produce outputs with reduced high-frequency noise.

(b) Representative trajectories of a tracked point showing that augmenting training labels with the LK-RSTC algorithm reduces model output jitter. The model outputs are high-pass filtered at 1.5 Hz to highlight the effect of label augmentation on jitter.

(c) Representative trajectories of a tracked point show that after applying the LK-RSTC filtering step, model outputs exhibit similar temporal jitter and tracking accuracy regardless of whether LK-RSTC label augmentation was used.

(d) RMSE between model trained with LK augmentation and ground truth (green), without LK augmentation and ground truth (magenta), and between the model with and without LK augmentation (cyan). Error is significantly lower for model trained with LK augmentation (paired t-test, $t_{35}=-3.598$, ***$p=0.00098$). The difference between the models with LK augmentation vs. without LK augmentation is significantly different from 0 (one-sided t-test, $t_{35}=15.833$, ***$p=1.573e-17$) after LK-RSTC filtering the outputs of both models. The dashed gray and black lines represent the average differences in tracked point position between the two approaches (with and without LK augmentation). The gray line (92 µm) shows the average difference when the human evaluator could not detect a difference between trajectories, while the black line (140 µm) shows the average difference when differences were detectable.

(e) Power spectral density analysis confirms that after LK-RSTC filtering, label augmentation has no impact on the frequency content of model outputs.

(f) In blind scoring by an ultrasound tracking expert, models with and without LK-RSTC label augmentation were perceptually similar after LK-RSTC filtering. The scorer showed no significant preference (binomial test, 127 vs. 109; statistic=0.538, $p = 0.268$, CI 0.472, 0.603).

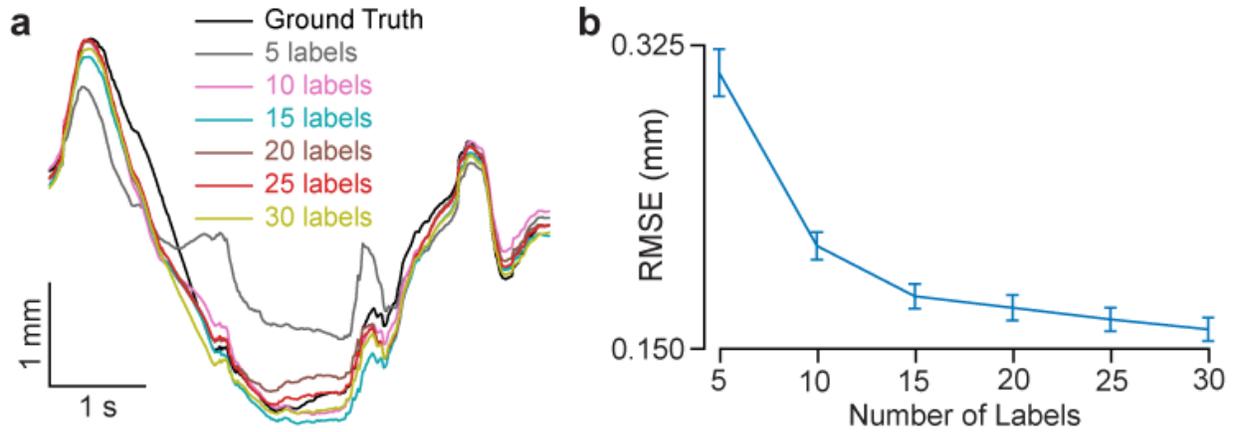

**Figure 6. DUSTrack tracking error decreases with the number of labeled frames.**

(A) Representative traces of tracked point position for increasing numbers of labeled frames, from 5 to 30.

(B) Root mean square error (RMSE) between the ground truth and DUSTrack output as a function of the number of labeled frames. The error decreases substantially between 5 and 15 labeled frames, with only marginal improvement observed between 15 and 30 labeled frames.

Error bars indicate S.E.M across 36 participants.

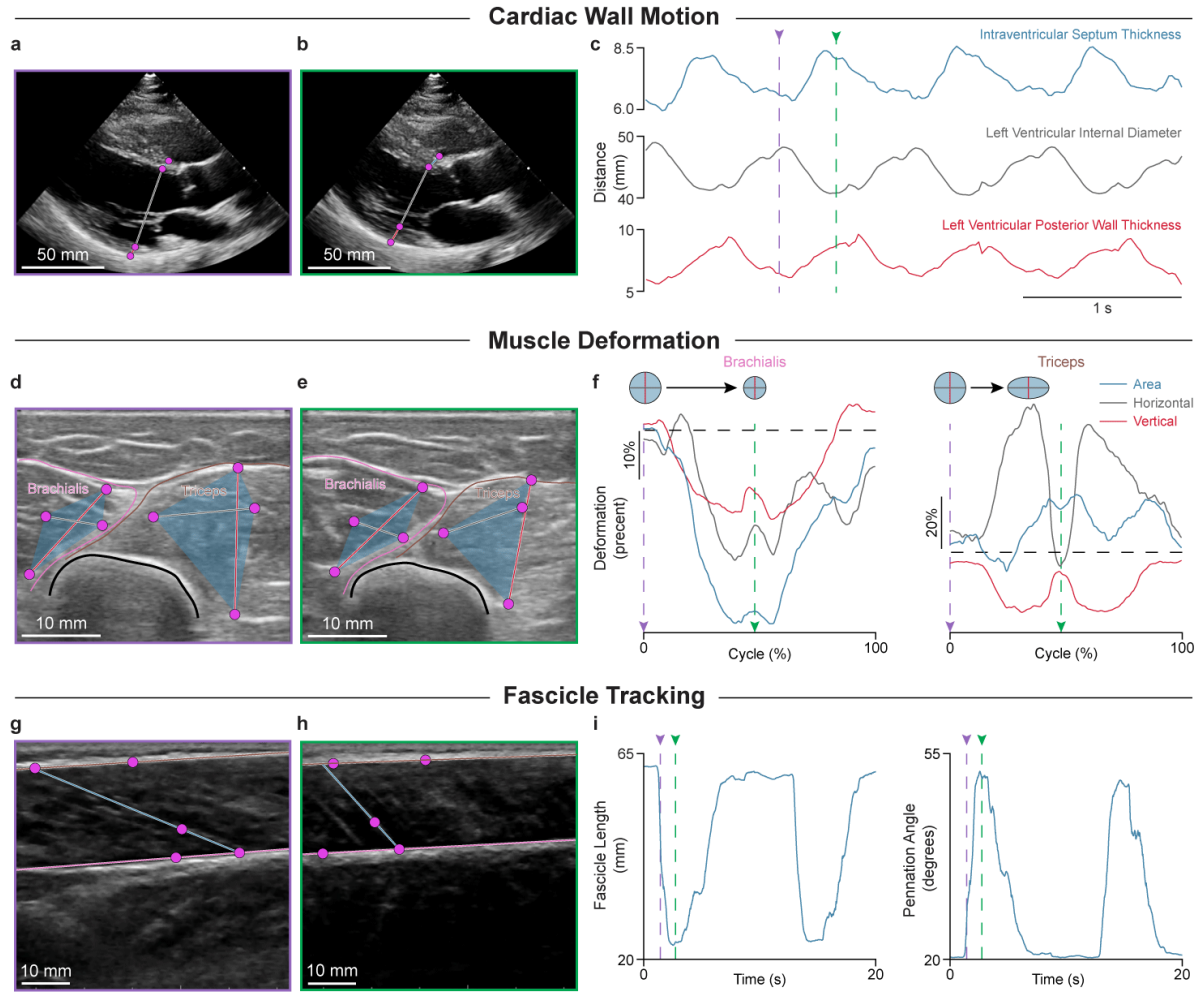

**Figure 7. Points tracked via DUSTrack enable extraction of clinically and biomechanically relevant measurements. Applications include cardiac wall motion tracking, muscle deformation tracking and fascicle tracking.**

(a-c) Using DUSTrack for cardiac wall motion tracking. Long-axis parasternal echocardiogram frames during systole (a) and diastole (b), showing tracked points (pink) used to measure cardiac structures: intraventricular septum thickness (blue), left ventricular internal diameter (gray), and left ventricular posterior wall thickness (red).

(c) Time series data showing the measured cardiac parameters across four cardiac cycles. Purple and blue dashed lines correspond to the frames shown in (a) and (b) respectively.

(d-f) Using DUSTrack for muscle deformation tracking. Transverse ultrasound images at the onset of extension (d) and retraction (e) of the upper arm during a reaching task, showing the brachialis and triceps muscles. Four tracked points per muscle are used to track distances within the muscle cross-section along superior-inferior (vertical, red) and medial-lateral (horizontal, gray) directions. The area enclosed by the four points is shown in blue.

(f) Muscle deformation patterns during one reaching cycle. The brachialis shows synchronized horizontal and vertical deformations, maintaining shape while reducing area. The triceps exhibits opposing deformations, indicating changing shape, and represented by the circle and oval sketches above. Purple and blue dashed lines correspond to the two frames in (d) and (e) at the onset of extension and retraction, respectively.

(g-i) Using DUSTrack for fascicle tracking. Medial gastrocnemius ultrasound images during the start (g) and end (h) of ankle plantarflexion. Points tracked on the superficial and deep aponeuroses, combined with fascicle orientation, allow calculation of fascicle length and pennation angle.

(i) Representative traces of fascicle length and pennation angle, derived from the tracked points. Purple and blue dashed lines correspond to the two frames in (g) and (h), respectively.

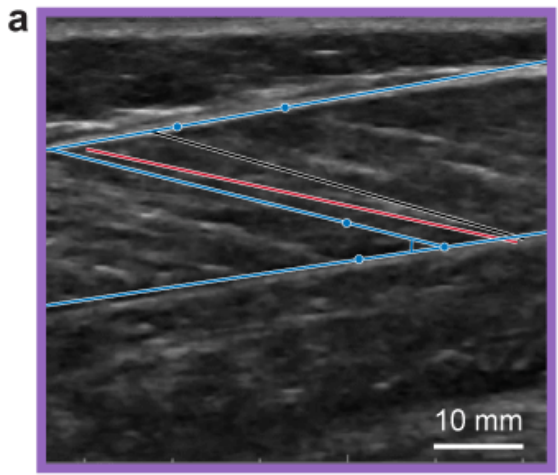 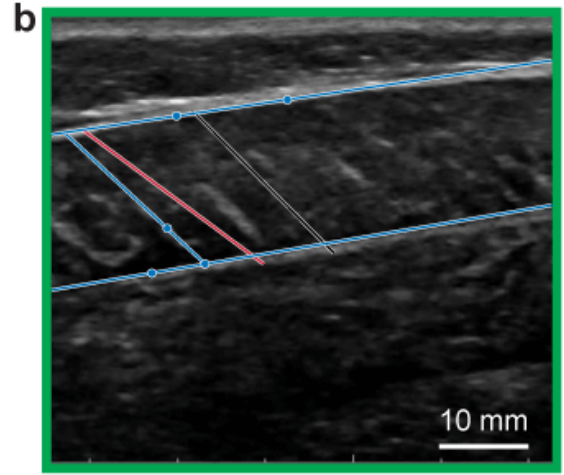

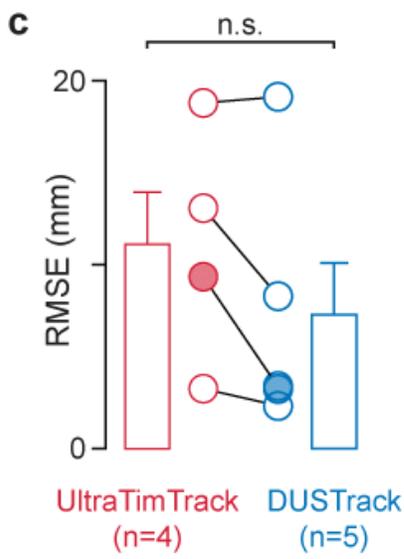 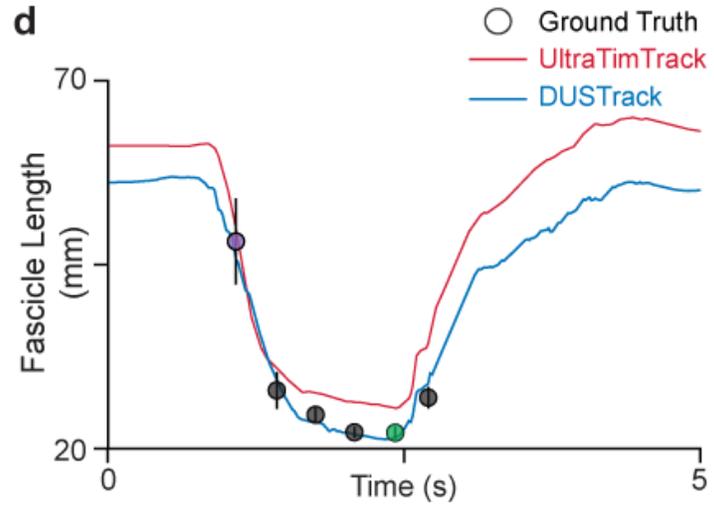

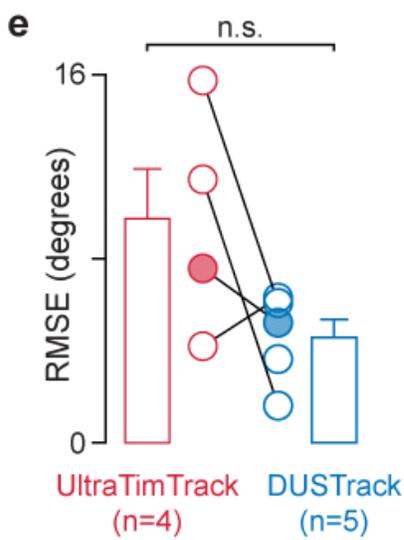 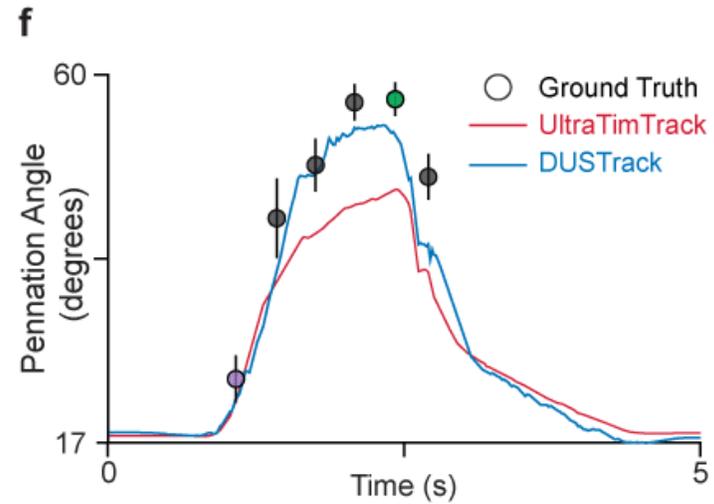

**Figure 8. Quantitative comparison of DUSTrack against UltraTimTrack demonstrates comparable accuracy in measuring fascicle length and pennation angle.**

(a-b) Ultrasound images of the medial gastrocnemius muscle during ankle plantarflexion show how we measure fascicle length and pennation angle. DUSTrack-tracked points (dots) define the upper and lower aponeuroses and fascicle orientation (blue lines). For comparison, we show UltraTimTrack's fascicle estimate (red line) and ground truth measurements (grey line). Fascicle length is measured between aponeurosis intersections, while pennation angle is measured between the fascicle and deep aponeurosis. Frames at the beginning (a) and end (b) of a plantarflexion motion are indicated by purple and green borders.

(c) Fascicle length measurements show similar accuracy between DUSTrack and UltraTimTrack, with no significant difference in root mean square error (RMSE) (paired t-test, $t_3$ = *1.88, p = 0.16*).

(d) Representative traces of fascicle length measurements across one plantarflexion cycle show how both methods track changes in length. Black circles indicate ground truth data from manual annotations, with error bars showing SEM across annotations. The purple and green filled circles correspond to the frames shown in (a) and (b).

(e) Pennation angle measurements also show comparable accuracy between methods, with DUSTrack showing a trend toward lower RMSE (paired t-test, $t_3$ = *1.66, p = 0.20*).

(f) Representative traces of pennation angle during one plantarflexion cycle demonstrate how both methods track angle variations. Gray circles show ground truth data from manual annotations, with error bars representing SEM across annotations. The purple and green filled circles correspond to the frames shown in (a) and (b).